\documentclass[10pt,twocolumn,letterpaper]{article}
\usepackage[dvipsnames]{xcolor}
\usepackage{iccv}
\usepackage{times}
\usepackage{epsfig}
\usepackage{graphicx}
\usepackage{amsmath}
\usepackage{amssymb}

\usepackage[pagebackref=true,breaklinks=true,letterpaper=true,colorlinks,bookmarks=false]{hyperref}

\usepackage{comment}
\usepackage{ifthen} 
\usepackage{overpic}
\usepackage{booktabs}
\usepackage{multirow}
\usepackage{rotating}
\usepackage{wrapfig}
\usepackage{amsmath}
\usepackage{amssymb}
\usepackage{bbm}
\usepackage{dsfont}
\usepackage{wrapfig}
\usepackage{verbatim} 
\usepackage{nicefrac} 
\usepackage{caption}
\usepackage{pifont}
\usepackage{colortbl} 
\usepackage{color}
\usepackage{xcolor}
\usepackage{balance}
\usepackage{marvosym}

\newcommand{\cmark}{\ding{51}}
\newcommand{\xmark}{\ding{55}}

\newcommand{\tg}[1]{\textcolor{LimeGreen}{#1}}

\newcommand{\ty}[1]{\textcolor{Goldenrod}{#1}}

\def\rowg{\rowcolor{gray!10}}
\def\cellg{\cellcolor{gray!10}}

\newcommand{\mathvec}[1]{\boldsymbol{#1}}
\newcommand{\mathmat}[1]{\mathbf{#1}}
\newcommand{\mathset}[1]{\mathcal{#1}}
\newcommand{\mathfuc}[1]{\mathtt{#1}}

\newcommand{\eqnref}[1]{Eq. \ref{#1}}
\newcommand{\secref}[1]{Sec. \ref{#1}}
\newcommand{\tabref}[1]{Tab. \ref{#1}}
\newcommand{\figref}[1]{Fig. \ref{#1}}

\def\ie{\emph{i.e.}}
\def\eg{\emph{e.g.}}
\def\etc{\emph{etc}}

\def\supp{\textit{\textcolor{BrickRed}{supplementary materials}}}

\iccvfinalcopy 

\def\iccvPaperID{1334} 

\ificcvfinal\fi

\begin{document}

\title{Advancing Referring Expression Segmentation Beyond Single Image}

\author{Yixuan Wu$^{*1}$ \quad Zhao Zhang$^{*2}$ \quad Chi Xie$^{3}$ \quad Feng Zhu\textsuperscript{\Letter}$^2$ \quad Rui Zhao$^2$\\$^1$Zhejiang University \quad $^2$SenseTime Research \quad  $^3$Tongji University\\
{\tt\small wyx\_chloe@zju.edu.cn \quad zzhang@mail.nankai.edu.cn \quad zhufeng@sensetime.com}
}

\maketitle
\ificcvfinal\thispagestyle{empty}\fi

\begin{abstract}
Referring Expression Segmentation (RES) is a widely explored multi-modal task, which endeavors to segment the pre-existing object within a single image with a given linguistic expression. 
However, in broader real-world scenarios, it is not always possible to determine if the described object exists in a specific image.
Typically, we have a collection of images, some of which may contain the described objects.
The current RES setting curbs its practicality in such situations.
To overcome this limitation, we propose a more realistic and general setting, named Group-wise Referring Expression Segmentation (GRES), which expands RES to a collection of related images, allowing the described objects to be present in a subset of input images.
To support this new setting, we introduce an elaborately compiled dataset named Grouped Referring Dataset (GRD), containing complete group-wise annotations of target objects described by given expressions. We also present a baseline method named Grouped Referring Segmenter (GRSer), which explicitly captures the language-vision and intra-group vision-vision interactions to achieve state-of-the-art results on the proposed GRES and related tasks, such as Co-Salient Object Detection  and RES.
Our dataset and codes will be publicly released in \href{https://github.com/yixuan730/group-res}{https://github.com/yixuan730/group-res}.
\end{abstract}

\section{Introduction}
\label{sec:intro}
\begin{figure}[t]
	\begin{center}
		\includegraphics[width=1.0\linewidth]{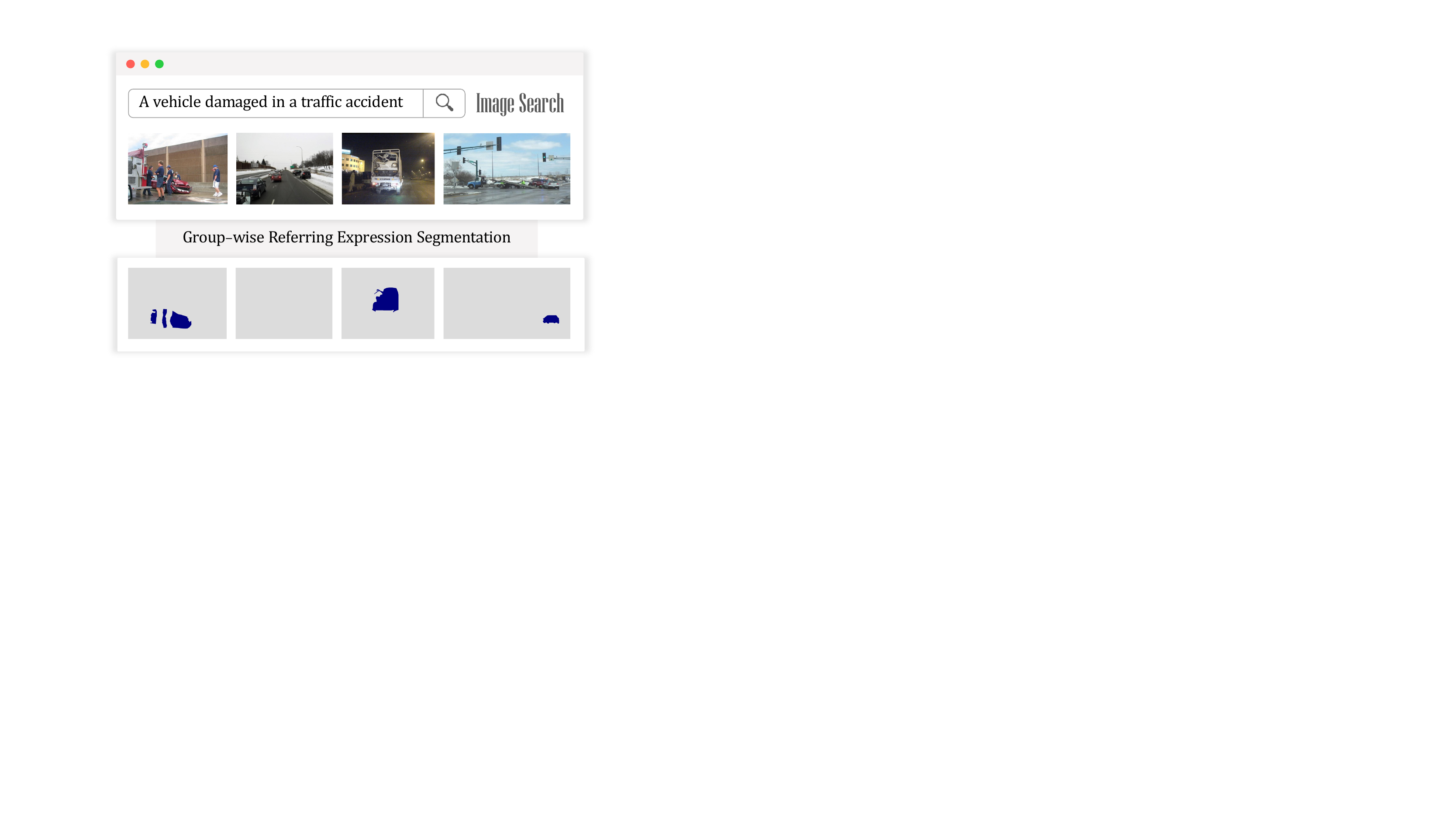}
	\end{center}
    \vspace{-15pt}
	\caption{
    \textbf{Real-world applications} of Group-wise Referring Expression Segmentation (GRES), which facilitates annotation auto-gathering from cluttered Internet images (upper),  multi-monitors joint inference (lower), \etc.
    } 
    \vspace{-11pt}
\label{fig:teaser}
\end{figure}

Segmenting target objects described by users in a collection of images is a fundamental but overlooked capability that facilitates various real-world applications (as illustrated in \figref{fig:teaser}), such as filtering and labeling cluttered internet images, multi-monitors event discovery, and mobile album retrieval. In recent years, Referring Expression Segmentation (RES) has become a research hotspot with great potentials to solve this demand. 
Various promising approaches~\cite{hu2016segmentation,wang2022cris,yang2022lavt,ding2022vlt,feng2021encoder} and datasets~\cite{kazemzadeh2014referitgame,yu2016modeling,mao2016generation,wu2020phrasecut} have contributed to significant advancements in this field.  
However, the setting of RES is overly idealistic. It aims to segment what has been known to exist in a single image described by a expression. This has restricted the practicality of RES in real-world situations, given that it is not always possible to determine if the described object exists in a specific image. Typically, we have a collection of images, some of which may contain the described objects.

To address this limitation, in this paper, we introduce a new realistic setting, namely Group-wise Referring Expression Segmentation (GRES), and define it as segmenting objects described in language expression from a group of related images. We establish the foundation of GRES in two aspects: 
firstly, a baseline method named Grouped Referring Segmenter (GRSer) that explicitly leverages language and intra-group vision connections to obtain promising results, and secondly, a meticulously annotated dataset, Group Referring Dataset (GRD), that ensures complete annotations of described objects across all images in a group.

Our proposed GRSer, illustrated in \figref{fig:pipeline}, facilitates a simultaneous processing of multiple input images with an expression, and generates segmentation masks for all described objects.
We devise a Triphasic Query Module (TQM), where the target objects not only queried by linguistic features, but also by intra-group visual features.
In contrast to segmenting based solely on linguistic expression, querying target objects with intra-group homo-modal visual features bridges the modal gap and assembles a more precise target concept.
In the proposed Heatmap Hierarchizer, these heatmaps generated by intra-group visual querying are ranked based on their confidences, and then jointly used to predict segmentation masks in condition of the ranking priorities.
Furthermore, we propose a mirror training strategy and triplet loss to learn anti-expression features, which are crucial for the TQM and Heatmap Hierarchizer, and enable GRSer to comprehend the image background and negative samples.
The promising performance of GRSer makes it a strong research baseline for GRES.

\begin{figure}[t]
\centering
     \begin{overpic}[grid=False,width=\columnwidth]{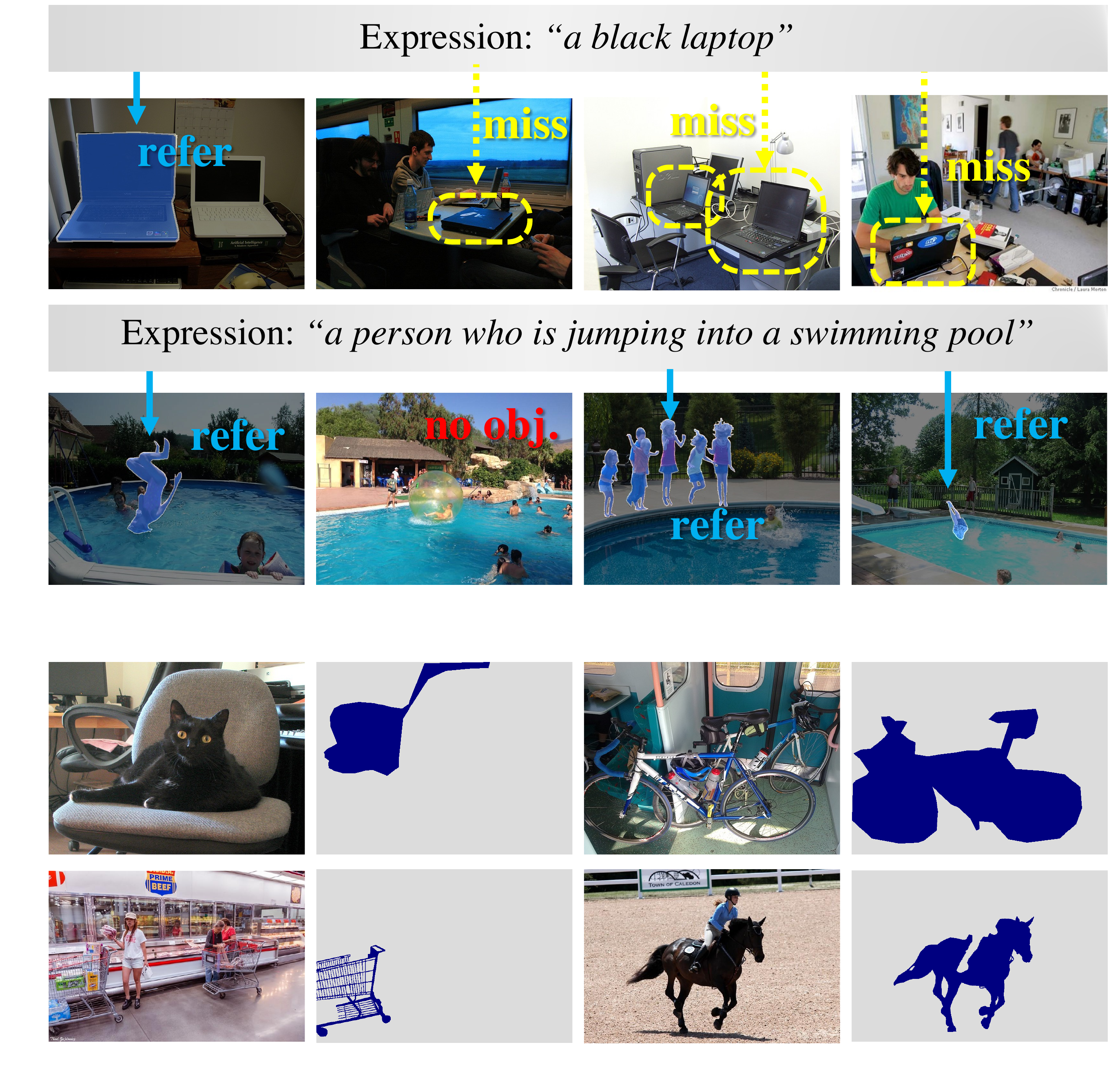}
     \put(0.1,20){\small{\rotatebox{90}{RefCOCOg}}}
     \put(0.1,7){\small{\rotatebox{90}{GRD}}}
     \put(32,40){\small{(a) Annotation completeness}}
     \put(0.1,70){\small{\rotatebox{90}{RefCOCOg}}}
     \put(0.1,50){\small{\rotatebox{90}{GRD}}}
     \put(36,-0.8){\small{(b) Annotation fineness}}
     \end{overpic} \\
    \vspace{-5pt}
   \caption{Proposed GRD \vs RefCOCOg on the annotation completeness and fineness.
   }
   \vspace{-17pt}
\label{fig:dataset}
\end{figure}
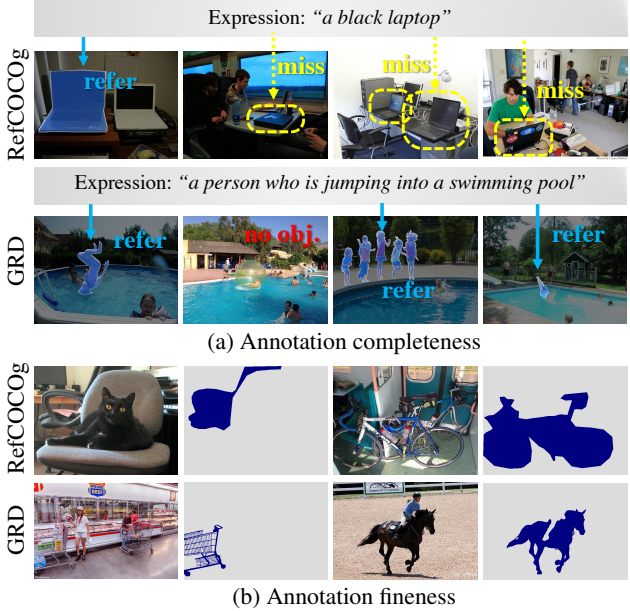
To facilitate the research in novel GRES setting, the GRD dataset is introduced, which effectively overcomes the incomplete annotation problem in current RES datasets~\cite{kazemzadeh2014referitgame,yu2016modeling,mao2016generation}.
For example, in \figref{fig:dataset}, RefCOCOg's expression of the 1st image also corresponds to objects in images 2, 3, and 4, but they are not annotated, causing erroneous false positive samples during evaluation if correctly segmented.
In contrast, expressions in GRD refers objects completely for all images across the dataset, including images without targets or with multiple targets.
Our GRD includes 16,480 positive object-expression pairs, and 41,231 reliable negative image-expression pairs.
Additionally, GRD collects images from Internet search engines by group keywords, where negative samples inherently exist in each group, making them hard negatives and effectively increasing the dataset's difficulty. 
Finally, as shown in \figref{fig:dataset}(b), compared with current RES datasets, GRD carefully labels details in segmentation masks, such as blocking and hollowing out, which contributes to a more accurate and reliable evaluation efficacy than existing datasets.

Our contributions can be summarized as:
\vspace{-4pt}
\begin{itemize}
    \item We formalize a Group-wise Referring Expression Segmentation (GRES) setting over the RES task, which advances user-specified object segmentation towards more practical applications.
    \vspace{-4pt}
    \item To support GRES research, 
    we present a meticulously compiled dataset named GRD, possessing complete group-wise annotations of target objects. The dataset will also benefit various other vision-language tasks.
    \vspace{-4pt}
    \item 
    Extensive experiments show the effectiveness and generality of the proposed baseline method, GRSer, which achieves SOTA results on the GRES and related tasks, such as Co-Salient Object Detection and RES.
\end{itemize}

\section{Related Work}
\label{sec:related_work}

\subsection{Referring Expression Segmentation (RES)}
\vspace{-5pt}
RES aims to ground the target object in the given image referred by the language and generate a corresponding segmentation mask.
\textbf{Methods.}
A common approach to solve RES is to first extract both vision and language features, and then fuse the multi-modal features to predict the mask. 
Early methods~\cite{hu2016segmentation,li2018referring,liu2017recurrent} simply concatenate visual features and language features extracted by convolutional neural networks (CNNs) and recurrent
neural networks (RNNs), respectively.
Due to the breakthrough of Transformer~\cite{wolf2020transformers,dosovitskiy2020image,liu2021swin}, a rich line of works begin to explore its remarkable fusion power for multi-modality. Some~\cite{yang2021bottom,liu2022instance,wang2022cris,feng2021encoder,ding2022vlt,kim2022restr,yang2022lavt} conduct cross-model alignment based on Transformer, others~\cite{yu2018mattnet,luo2020cascade,liu2021cross,luo2020multi,yang2021bottom} adopt various attention mechanisms to achieve better feature weighting and fusing. 
There are some works to explore how to solve RES working with related tasks, such as visual grounding~\cite{luo2020multi,li2021referring,zhu2022seqtr,liu2023polyformer}, zero/one-shot segmentation~\cite{luddecke2022CLIPseg}, interactive segmentation~\cite{ding2020phraseclick}, unified segmentation~\cite{zou2022xdecoder}, and referring expression generation~\cite{huang2022unified}.
\textbf{Datasets.}
Several datasets have been introduced to evaluate the performance of RES methods, including RefClef~\cite{kazemzadeh2014referitgame}, RefCOCO~\cite{yu2016modeling}, RefCOCO+~\cite{yu2016modeling}, RefCOCOg (G-Ref)~\cite{mao2016generation}, and PhraseCut~\cite{wu2020phrasecut}. RefClef, RefCOCO, and RefCOCO+ are collected interactively in a two-player game, named ReferitGame~\cite{kazemzadeh2014referitgame}, thus the given expressions are more concise and less flowery. 
Among them, RefCOCO+ bans location words in expressions, making it more challenging.
RefCOCOg is collected non-interactively, resulting in more complex expressions, often full sentences instead of phrases.
PhraseCut's phases, consist of attribute, category, and relationship, are automatically generated by predefined templates and existing annotations from Visual Genome~\cite{krishna2017visualgenome}. 
The above datasets fail to serve as reliable evaluation datasets for GRES setting due to their image-text pairs are one-to-one matched, which leads to
incomplete annotation for target objects in unmatched images. More datasets comparison can be found in \tabref{tab:data_comp}.

\subsection{Co-Salient Object Detection (Co-SOD)}
Co-SOD is a recent research focus \cite{zhang2019csmg,zhang2020CoADNet,Jin2020ICNet,zhang2020gicd,zhang2021deepacg,Zhang2020GCAGC,fan2021GCoNet,zhang2021summarize,deng2021cosod3k,yu2022democracy,zhu2023corp}, aiming to discover the common semantic objects in a group of related images. 
In this task, the target object does not need to be specified by language expression, while required to appear commonly in all images. Co-SOD methods need to perceive what the common objects are from the pure visual modality, and then segment them.
Historically, researchers refer to Co-SOD as ``detection", but its outputs are actually segmentation maps.
\textbf{Methods.}
Recently, many impressive Co-SOD methods have arisen, focusing primarily on obtaining co-representations of common objects to guide target object segmentation.
Co-representations can be obtained through methods like feature concatenation~\cite{wei2019deep}, linear addition~\cite{zhang2020gicd}, channel shuffling~\cite{zhang2020CoADNet}, graph neural networks~\cite{jiang2019unified,Zhang2020GCAGC}, and iterative purification~\cite{zhu2023corp}.
There is also a body of research work focused on intra-group information exchange,
such as using pair-wise similarity map~\cite{Jin2020ICNet}, dynamic convolution~\cite{zhang2021summarize}, group affinity~\cite{fan2021GCoNet}, and transformers~\cite{ge2022tcnet,su2022unified}.
Moreover, besides these central lines of exploring, efforts have been made to enhance the Co-SOD model through data enhancement~\cite{zhang2020gicd,zhang2021summarize}, training strategies~\cite{fan2021GCoNet,yu2022democracy}, adversarial attack preventing~\cite{gao2022can}.
\textbf{Datasets.}
Co-SOD datasets include iCoseg~\cite{batra2010icoseg}, MSRC~\cite{winn2005object}, CoSal2015~\cite{zhang2016detection}, CoSOD3k~\cite{deng2021cosod3k}, and CoCA~\cite{zhang2020gicd}. Early datasets such as iCoseg and MSRC contain co-salient objects with similar appearance in similar scenes. CoSal2015 and CoSOD3k are large-scale datasets, featuring target objects with varying appearance in the same category. CoCA, the latest dataset, presents a more challenging setting with at least one extraneous salient object in each image, requiring the model to identify the target object in cluttered scenes.
Although the data sets have favorable grouping scenarios, they lack expressions and negative samples, making them unsuitable for direct use as evaluation dataset for GRES.

\begin{figure*}
\begin{center}
\includegraphics[width=0.98\linewidth]{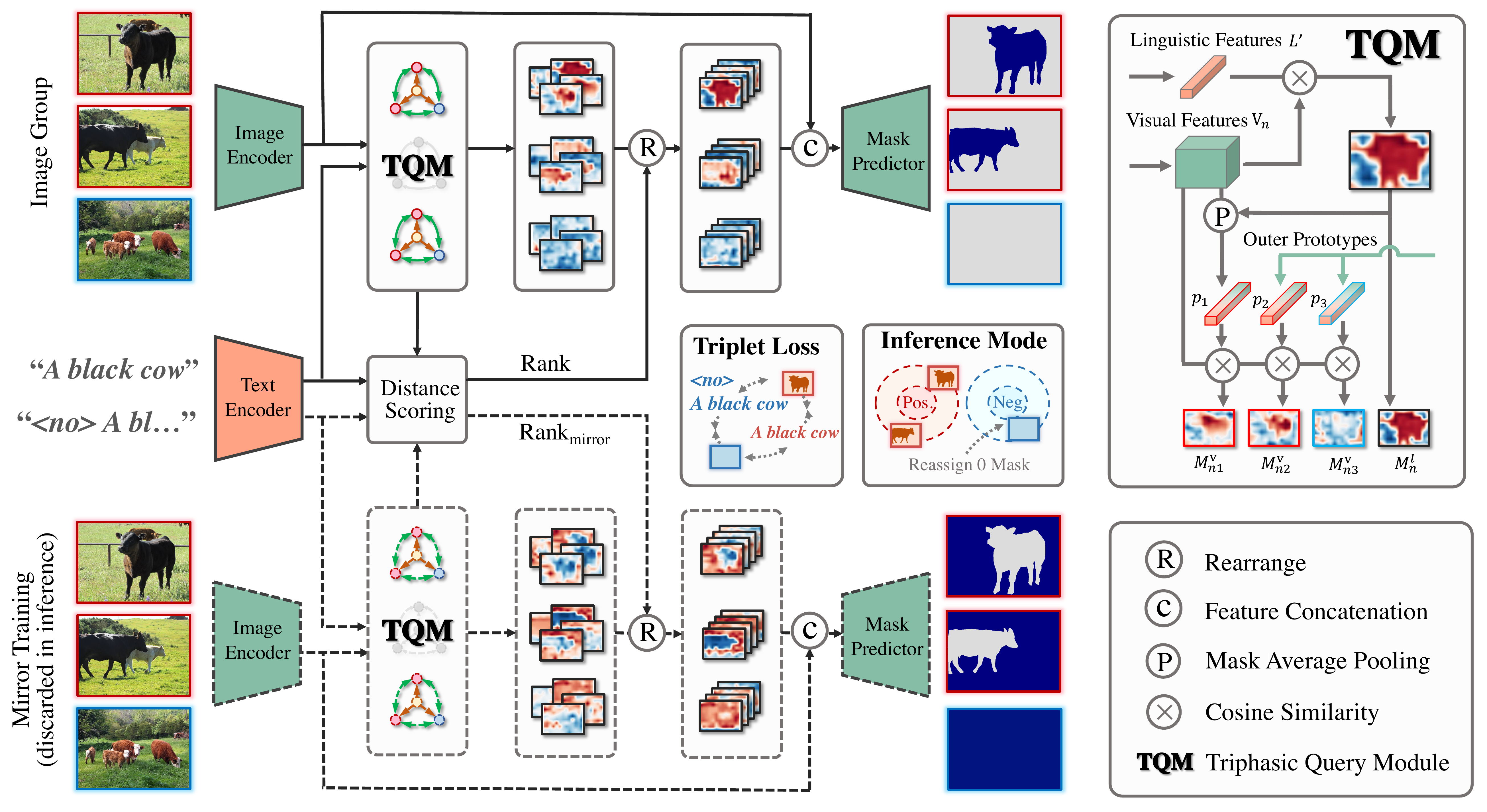}
\end{center}
    \vspace{-15pt}
   \caption{\textbf{The pipeline of proposed GRSer}. First, grouped input images, given an expression together with its anti-expression ($<$no$>$ prefix added), are encoded by image and text encoder, respectively, and fed into a triphasic query module (TQM), to generate a set of heatmaps that indicate the most discriminating region in the visual feature map responding to the target object.
   Next, these heatmaps are rearranged according to their correlation with the description, and then concatenated with visual features for mask prediction.
   In training, triplet loss and segmentation loss are both applied, and a mirror training strategy (dotted line) is introduced to better comprehend the anti-expression and image background.   
   In inference, the mirror training will be discarded, and images close to the anti-expression are reassigned $0$ masks.}
       \vspace{-13pt}
\label{fig:pipeline}
\end{figure*}

\section{Proposed Method}
\label{sec:method}
\subsection{Overview}
The pipeline of our Grouped Referring Segmenter (GRSer) is demonstrated in Fig.~\ref{fig:pipeline}. Given an expression that specifies an object, a group of related images are processed simultaneously, and then all corresponding pixel-wise masks of the target object are output. In particular, for the negative sample (\ie, image without target object), its output mask is $0$ mask. There are four modules in our GRSer, including a multi-modal encoder, a triphasic query module, a heatmap hierarchizer, and a mask predictor.

\noindent \textbf{Text \& Image Encoder.} BERT~\cite{devlin-etal-2019-bert} is employed to embed the expression into linguistic features $\mathmat{L} \in \mathbb{R}^{C_{l}}$, where $C_{l}$ is the number of channels for the language feature. Meanwhile, we construct an anti-expression by adding a prefix $<$no$>$ to the given expression, which is embedded as linguistic anti-features $\mathmat{L}^{anti} \in \mathbb{R}^{C_{l}}$. We follow LAVT~\cite{yang2022lavt} to perform visual encoding to obtain visual features $\mathmat{V}_{n} \in \mathbb{R}^{C_{v}\times H\times W}$ for each image $x_{n}$ in the group ($n=1,\ldots,N$), where $N$ is the number of images in one group, and $C_{v}$, $H$, and $W$ denote the channel number, height, and width, respectively. For more details about the encoder and decoder, please refer to \supp.

\noindent \textbf{TQM \& Heatmap Hierarchizer.} The language-vision and intra-group vision-vision semantic relations are explicitly captured in proposed TQM (\secref{sec:GCoQuery}) to produce heatmaps, which reflect the spatial relation between linguistic and intra-group visual features. 
And these heatmaps are ranked and rearranged in heatmap hierarchizer (\secref{sec:HMapRank}) according to their importance with the expression to better activate their locating capability for mask prediction.

\noindent \textbf{Mask Predictor.} The well-ranked heatmaps are concatenated with visual features $\mathmat{V}_{n}$ to obtain the triphasic features $\mathmat{z}_{n}$, which integrates the discriminative cues of target object in TQM and heatmap hierarchizer. 
$\mathmat{z}_{n}$ is used to distinguish positive or negative samples, and predict the segmentation masks. 
\textbf{In inference}, 
the positive distance $d^{pos}\!=\!\mathfuc{d}(\mathmat{z}_{n},\mathmat{L})$ and negative distance $d^{neg}\!=\!\mathfuc{d}(\mathmat{z}_{n},\mathmat{L}^{anti})$ are computed, where the Euclidean Distance $\mathfuc{d(\cdot)}$ is applied.
If $d^{pos}+ m<d^{neg}$ ($m$ is the margin value), then image $x_{n}$ is recognized as a positive sample, and its  $\mathmat{z}_{n}$ is then transmitted to the decoder to output segmentation mask. If not, $0$ mask is reassigned as negative output.

\vspace{-3pt}
\subsection{Triphasic Query Module (TQM)}
\vspace{-1pt}
\label{sec:GCoQuery}

\begin{figure*}
\begin{center}
\includegraphics[width=0.90\linewidth]{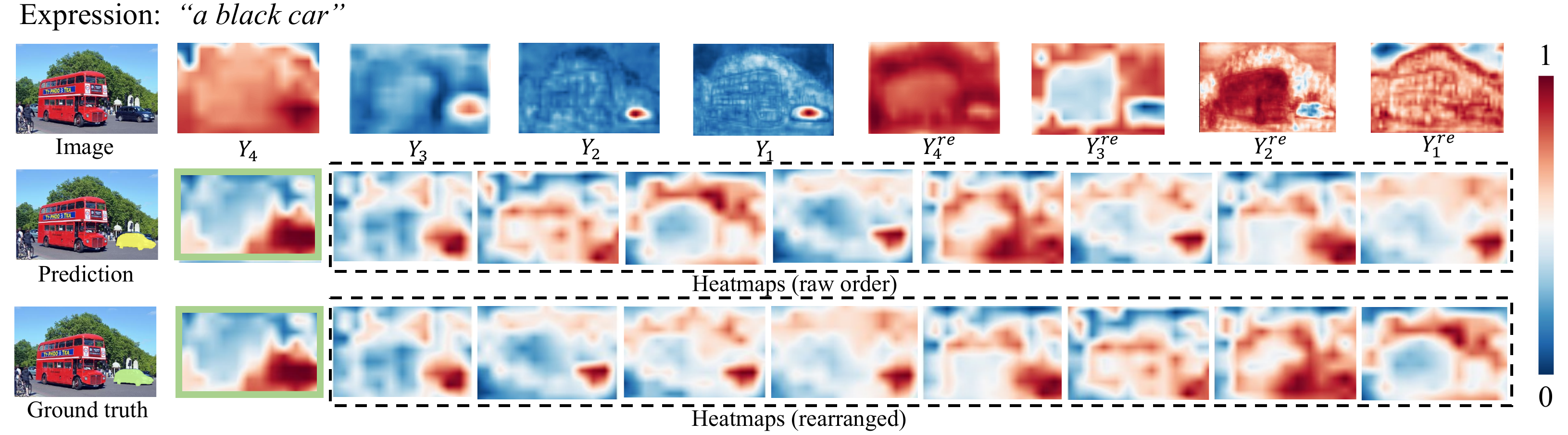}
\end{center}
\vskip -1.7 em
   \caption{\textbf{Visualizations of the prediction mask, feature maps, and heatmaps} on an example from the G-RefCg test set. The leftmost column demonstrates the input image, predicted mask (in \ty{yellow}) and ground-truth mask (in \tg{green}). In the first row of other columns, we visualize the feature maps in decoder (\ie, $\mathmat{Y}_{i}$) activated by linguistic feature $\mathmat{L}$, and feature maps in decoder (\ie, $\mathmat{Y}_{i}^{anti}$) activated by linguistic anti-feature $\mathmat{L}^{anti}$ in our proposed mirror training process (\secref{sec:loss}). In the second and third row, we demonstrate the language-activated heatmap $\mathmat{M}^{l}$ (the one inside the solid \tg{green} border) and vision-activated heatmaps $\mathmat{M}^{v}$ (the others inside the dotted border) . Note that the raw order of heatmaps are shown in the second row, and the well-ranked heatmaps are demonstrated in the third row. Best viewed in color.}
       \vspace{-13pt}
\label{fig:vis}
\end{figure*}

Due to the inherent modality gap, directly querying objects through linguistic features often results in rougher language-activated heatmaps (\eg, the 2nd image in the bottom row of \figref{fig:vis}).
We resort to intra-group homo-modal visual features to act as ``experts", offering suggested heatmaps from their perspectives.
To this end, we devise the TQM, where ``triphasic" means that the target object not only queried by linguistic features, but also by intra-group homo-modal visual features.

In the right top of \figref{fig:pipeline}, we take one image $x_{n}$ as an example to illustrate the detailed process. 
First, in order to detect the most discriminating region in the visual feature map responded to the referring expression, a language-activated heatmap $\mathmat{M}^{l}_{n}\in \mathbb{R}^{H\times W}$ is generated. Specifically, the cosine similarity is computed between the flattened visual features $\mathmat{V}_{n} \in \mathbb{R}^{C_{v}\times HW}$ and linguistic features $\mathmat{L}'\!=\!\omega_{l}(\mathmat{L}) \in \mathbb{R}^{C_{v}}$, where a $1\times1$ convolution layer $\omega_{l}$ with $C_{v}$ number of output channels are deployed to align the cross-modal features. This is denoted as
\begin{equation}
    {\mathmat{M}^{l}_{n}=\frac{\mathmat{V}^{T}_{n}\cdot \mathmat{L}'}{\Vert \mathmat{V}_{n}\Vert \,\Vert \mathmat{L}' \Vert}}.
\end{equation}
Second, $\mathmat{M}^{l}_{n}$ is element-wise multiplied with visual features $\mathmat{V}_{n}$, and the output features are averaged along spatial dimension (\ie, $H\times W$) with mask average pooling to generate a prototype $\mathvec{p}_{n}\!\in\!\mathbb{R}^{C_{v}}$ corresponding to image $x_{n}$, as
\begin{equation}
    {\mathvec{p}_{n}=\mathfuc{avg}(\mathmat{M}^{l}_{n}\odot \mathmat{V}_{n})},
\end{equation}
where $\mathmat{M}^{l}_{n}$ is broadcast to the same size as $\mathmat{V}_{n}$, and $\odot$ denotes the element-wise multiplication. 
In this manner, a group of prototypes $\{\mathvec{p}_{i}\}_{i=1}^{N}$ is generated, with each prototype corresponding to one image from a group. 
Intuitively, the prototype integrates visual features of the target object.

Next, the intra-group queries are conducted between current image $x_{n}$ and a group of prototypes, and these prototypes serve as ``experts" to provide localization heatmap suggestions from their perspectives.

In details, the cosine similarity is computed between the flattened visual features $\mathmat{V}_{n}$ and each prototype $\mathvec{p}_{i}$ from $\{\mathvec{p}_{i}\}_{i=1}^{N}$ one-by-one, and then produce $N$ vision-activated heatmaps $\mathmat{M}^{v}_{n}=\{\mathmat{M}^{v}_{ni}\}_{i=1}^{N}$, as 

\begin{equation}
    {\mathmat{M}^{v}_{ni}=\frac{\mathmat{V}^{T}_{n}\cdot \mathvec{p}_{i}}{\Vert \mathmat{V}_{n}\Vert \,\Vert \mathvec{p}_{i} \Vert}},
\end{equation}
where $n$ denotes the index of image in a group, and $i$ denotes the index of prototype in a group. As shown in \figref{fig:vis}, these four $\mathmat{M}^{v}_{ni}$ (the 3rd - 6th in the bottom row) show stronger locating capability than the $\mathmat{M}^{l}_{n}$ (the 2nd in the bottom row), which thus provide more accurate guidance for mask prediction.

\subsection{Heatmap Hierarchizer}\label{sec:HMapRank}
Considering that the vision-activated heatmaps suggested by ``experts" from TQM can be uneven, especially when there are negative samples. 
For example, in \figref{fig:vis}, 
prototypes come from negative samples tend to generate counterfactual localization heatmaps (the 7th - 10th in the bottom row).
We need experts to give confidence of their suggestions to determine the heatmap priority in following prediction.
To this end, we propose a heatmap hierarchizer to rank and rearrange these vision-activated heatmaps based on a confidence evaluation strategy.

To get the rank of different heatmaps, we define a scoring criterion based on the multi-modal representation distance. Specifically, we compute the Euclidean Distance~\cite{dokmanic2015euclidean} between each prototype from the group $\{\mathvec{p}_{i}\}_{i=1}^{N}$ and linguistic features $\mathmat{L}$ to get the positive score $\{s^{pos}_{i}\}_{i=1}^{N}$. Negative score $\{s^{neg}_{i}\}_{i=1}^{N}$ is also obtained by computing Euclidean Distance between $\{\mathvec{p}_{i}\}_{i=1}^{N}$ and linguistic anti-features $\mathmat{L}^{anti}$. In this way, a smaller $s^{pos}_{i}$ indicates that the prototype $\mathvec{p}_{i}$ gets closer to the target object, which means its corresponding generated heatmap $\mathmat{M}_{ni}^{v}$ is more reliable. Inversely, a smaller $s^{neg}_{i}$ indicates the prototype $\mathvec{p}_{i}$ fits the background (\ie, outside of the target object in a image) better. Then, we obtain the positive rank $R^{pos}$ and negative rank $R^{neg}$ for $N$ vision-activated heatmaps $\mathmat{M}^{v}_{n}=\{\mathmat{M}^{v}_{ni}\}_{i=1}^{N}$, according to corresponding positive score $s^{pos}_{i}$ (from smallest to largest) and negative score $s^{neg}_{i}$ (from largest to smallest), respectively. The positive rank $R^{pos}$ and negative rank $R^{neg}$ are summed as the final rank to rearrange $\mathmat{M}^{v}_{n}$ by
\begin{equation}
    {\overline{\mathmat{M}}_{n}^{v}=\mathfuc{rearrange}\left(\mathmat{M}^{v}_{n}|R^{pos}+R^{neg}\right)},
\end{equation}
where $\mathfuc{rearrange}(\cdot)$ means changing the channel-wise order of these stacked heatmaps.
These heatmaps are then concatenated with visual features $\mathmat{V}_{n}$ to get triphasic features $\mathmat{z}_{n}$ for mask prediction. In \figref{fig:vis}, it can be seen that heatmaps with lower confidence (generated by negative samples) are relegated to the back after rearrangement. 

\subsection{Training Objectives}\label{sec:loss}
\noindent \textbf{Training with Negative Samples.}
For training, we set the ratio between positive samples $x^{pos}$ (\ie, image containing target object referred by the expression) and negative samples $x^{neg}$ (\ie, noisy image where no target object exists) in each image group as $1:1$. The training objectives are twofold: (1) Triplet margin loss to empower model with recognition ability for negative samples; (2) Cross-entropy loss to optimize the model's segmentation performance.

\noindent \textbf{Triplet Margin Loss.}
The goal of triplet margin loss~\cite{dong2018triplet} is to bring closer together the anchor and the positive example, while pull the anchor from the negative example away, as is illustrated in \eqnref{eq:trils}. The Euclidean Distance $\mathfuc{d(\cdot)}$ is applied, and $m$ is the margin value. For a positive sample $x^{pos}$, its triphasic features $\mathmat{z}_{n}$ is regarded as the anchor, and linguistic features $\mathmat{L}$ and anti-features $\mathmat{L}^{anti}$ are regarded as the positive and negative examples, respectively. And for a negative sample $x^{neg}$, $\mathmat{L}^{anti}$ and $\mathmat{L}$ are regarded as its positive and negative examples instead. The triplet margin loss is computed as
\begin{equation}\label{eq:trils}
    \resizebox{.9\linewidth}{!}{$
    \displaystyle{
\mathset{L}_{tri}\!=\!
\begin{cases}
\mathfuc{max}\!\left(\mathfuc{d}(\mathmat{z}_{n},\mathmat{L})\!-\!\mathfuc{d}(\mathmat{z}_{n},\mathmat{L}^{anti})\!+\!m,0\right) 
\ \ \text{for} \ x^{pos} \! \\
\mathfuc{max}\!\left(\mathfuc{d}(\mathmat{z}_{n},\mathmat{L}^{anti})\!-\!\mathfuc{d}(\mathmat{z}_{n},\mathmat{L})\!+\!m,0\right)
\ \ \text{for} \ x^{neg}  \!
\end{cases} }$}
\end{equation}

\noindent \textbf{Mirror Training Strategy.}
To further force our model to comprehend the semantics contained in linguistic anti-features $\mathmat{L}^{anti}$, we design a mirror training strategy. Intuitively, linguistic anti-features represent the opposite semantics of the given expression, and thus we explicitly relate the linguistic anti-features to the image background (\ie, outside of the target object in an image). Specifically, during training, on the basis of original pipeline, we add an additional mirror one that swaps the roles of $\mathmat{L}$ and $\mathmat{L}^{anti}$, and corresponding ground-truth mask is replaced with the background (\ie, $1-\mathmat{Y}$, where $\mathmat{Y}$ denotes the ground-truth mask for the target object). As shown in the first row of \figref{fig:vis}, the feature maps (\ie, $\mathmat{Y}^{anti}_{i}$) in decoder activated by $\mathmat{L}^{anti}$ exactly focus on the background outside of the target object. The cross-entropy loss is applied for mirror training, denoted as $\mathset{L}^{mirr}_{ce}$.

\noindent \textbf{Objective Function.}
Note that only positive samples $x^{pos}$ are included for computing cross-entropy loss, while all samples (\ie, $x^{pos}$ and $x^{neg}$) are used for computing triplet margin loss. We adopt the increasing weighting strategy for triplet margin loss to optimize the training process, by
\begin{equation}
    {\mathset{L}=\mathset{L}_{ce}(\hat{\mathmat{Y}},\mathmat{Y}) + \lambda\mathset{L}^{mirr}_{ce}(\hat{\mathmat{Y}}^{anti},1-\mathmat{Y}) + \frac{t}{T}\mathset{L}_{tri}},
\end{equation}
where $t$ and $T$ denote the current training epoch and total number of training epochs, respectively; $\lambda$ is a hyper-parameter to weigh the importance of mirror training strategy; $\hat{\mathmat{Y}}$ denotes predicted mask referred by the expression, and $\hat{\mathmat{Y}}^{anti}$ denotes predicted mask referred by the anti-expression obtained in mirror training strategy.  
\vspace{-8pt}

\section{Proposed Dataset}
\label{sec:dataset}
\subsection{Dataset Highlights}
\vspace{-1pt}
Within a collection of images, for a given expression, we label all described objects in all images without any omission.
This constitutes the fundamental attribute that distinguishes GRD from its counterparts.
For instance, in RefCOCOg's two samples shown in \figref{fig:dataset}, the first image's expression is ``man in blue clothes", while the same object in the second image lacks annotation. 
This flaw renders the expression valid only in one image, making other images in the dataset unsuitable as negative samples.
In addition to complete annotation, there are some features that make GRD exceptional.
One is that the images in each group of GRD are related, so even if the described target does not appear on some images in the group, the scenes in these images are often close to the description, which makes this dataset more challenging. 
Additionally, our delicate annotation in \figref{fig:dataset} enables objective evaluation of model performance compared to current RES datasets. 
More features can be found in \tabref{tab:data_comp}.
Thanks to these features, GRD can help many other vision-language tasks, such as visual grounding, RES, and grounding caption. 
GRD is freely available for non-commercial research purposes.

\begin{table}[t]
\centering
\renewcommand{\arraystretch}{1.2}
\renewcommand{\tabcolsep}{1.02mm}
\caption{\textbf{Valuable features bring by GRD dataset.}
``\textit{scene grouping}" means samples are grouped by similar scenes.
``\textit{complete annotation}" means any object satisfying the given description is annotated across dataset. 
In this case, samples without the label for a specific expression could be reliably considered as ``\textit{certified negative samples}" for this expression. 
If there are ``\textit{multiple referred objects}" described in an image, all of them are annotated without omission.
``\textit{meticulous masks}" are provided to fits the object perfectly, especially for the hollowed-out and blocking areas. 
``\textit{object-centric}" means the dataset concentrates on objects rather than broad concepts like grass and sky. ``\textit{avg. expression length}" represents the average expression length. 
RC, RC+, RCg, RCF, and PC denote RefCOCO, RefCOCO+, RefCOCOg, RefClef, and PhaseCut, respectively.
}
    \vspace{-8pt}
\begin{tabular}{l|ccccc|c}
\hline
  & RC & RC+ & RCg & RCF & PC & GRD  \\ \hline
\rowg scene grouping & \xmark  & \xmark & \xmark & \xmark & \xmark & \cmark \\
complete annotation & \xmark  & \xmark & \xmark & \xmark & \xmark & \cmark \\
   \rowg  certified neg. samples & \xmark  & \xmark & \xmark & \xmark & \xmark & \cmark \\
   multi. referred objects & \xmark  & \xmark & \xmark & \xmark & \cmark & \cmark \\
 \rowg meticulous masks & \xmark  & \xmark & \xmark & \xmark & \xmark & \cmark \\
   object-centric & \cmark  & \cmark & \cmark & \xmark & \cmark & \cmark \\ 
  \rowg avg. expression length & 3.6  & 3.5  & 8.4  & 3.5 & 2.0 & 5.9 \\ 
  \hline
\end{tabular}
    \vspace{-14pt}
\label{tab:data_comp}
\end{table}

\subsection{Construction Procedures}
\vspace{-1pt}
We collect images searching from Flickr\footnote{https://www.flickr.com}.
If crawling directly according to the expression, we usually get the iconic images,
which appear in profile, unobstructed near the center of a neatly composed photo. 
In order to meet the real situation and increase the challenge, we employ the combination of target keywords and scene keywords to crawl a group of related images from search engines. 
Consequently, the images involves intricate scenes, \ie, non-iconic images~\cite{lin2014microsoft}. 
Then, for each group of images, we carefully propose several related expressions to be annotated. 
The announcers will segment the objects in the group according to these expressions, without excluding any referred objects.
This completeness allows our dataset to accurately assess the model's performance on negative samples.
Each object annotation takes an average of 3 minutes to precisely define edges and remove hollow areas, guaranteeing accurate evaluation of model segmentation performances.

\subsection{Datset Statistics.}
\vspace{-1pt}

The GRD dataset contains 10,578 images.
It includes 106 scenes (groups), such as indoor, outdoor and sports ground.
Each group has around 100 images and 3 well-designed expressions referring to various number of positive and negative samples.
In total, the dataset is annotated with 316 expressions, resulting in 31,524 positive or negative image-text pairs.
The expressions have an average length of 5.9 words.
More statistics and examples can be viewed in \supp.

\begin{table*}[t!]
\centering
\renewcommand{\arraystretch}{1}
\renewcommand{\tabcolsep}{1.7mm}
\caption{
		\textbf{Quantitative comparisons with RES methods} in terms of mean Intersection-over-Union ($\text{mIoU}$) for the RES setting and our proposed $\overline{\text{mIoU}}$ for the GRES setting on the G-RefC, G-RefC+, G-RefCg, and our proposed GRD datasets. The best results are marked in \textbf{bold}.
	}
     \vspace{-8pt}
\begin{tabular}{l|c|cccc|cccc}
\hline 
\multirow{2}{*}{Method} & \multirow{2}{*}{Pub} & \multicolumn{4}{c|}{GRES (with negative samples)}                                                                                            & \multicolumn{4}{c}{RES (no negative samples)}                                                                                          \\ 
                        &                      & G-RefC & G-RefC+ & G-RefCg &GRD & \multicolumn{1}{c}{G-RefC} & \multicolumn{1}{c}{G-RefC+} & \multicolumn{1}{c}{G-RefCg} & \multicolumn{1}{c}{GRD} \\ 
                        \hline 
                \rowg EFN &\scriptsize CVPR21 \cite{feng2021encoder} &25.42&22.32&20.77&15.29&63.52&55.37&52.88&31.57 \\    
                    VLT &\scriptsize PAMI22 \cite{ding2022vlt} &26.87&24.38&22.83&16.58&66.98&59.14&51.73&33.78 \\ 
                \rowg CRIS &\scriptsize CVPR22 \cite{wang2022cris}  &29.31&27.27&24.74&19.33&70.62&68.12&58.93&41.23 \\ 
                      LAVT &\scriptsize CVPR22 \cite{yang2022lavt}  &30.22&27.14&24.38&18.48&75.27&67.93&59.94&39.14 \\ 
                        \hline
                 \rowg GRSer   & Ours &\textbf{84.77}&\textbf{78.44}&\textbf{75.32}&\textbf{57.12} &\textbf{79.33}&\textbf{70.38}&\textbf{65.47}&\textbf{47.25}\\ 
                         \hline

\end{tabular}
     \vspace{-7pt}

\label{tab:res}
\end{table*}

\begin{table*}[t!]
	\centering
	\renewcommand{\arraystretch}{1.1}
	\renewcommand{\tabcolsep}{1.48mm}
	\caption{
		\textbf{Quantitative comparisons with Co-SOD methods} in terms of mean absolute error (MAE)\cite{cheng2013efficient},
		maximum F-measure~\cite{borji2015SalObjBenchmark} ($F_{\max}$),
		S-measure~\cite{fan2017structure} ($S_{\alpha}$),
		and mean E-measure~\cite{Fan2018Em} ($E_{\xi}$) on the CoCA~\cite{zhang2020gicd} dataset. 
		``$\uparrow$'' means that the higher the numerical value, the better the model performance, and vice versa for ``$\downarrow$''. The best results are marked in \textbf{bold}.
	}
     \vspace{-8pt}
 
	\begin{tabular}{lc|ccccccccc|c}
		\hline 
		& &  CSMG & GCAGC  &  GICD   & ICNet & CoEG & DeepACG & GCoNet & CADC  & CoRP  & GRSer  \\
		[-0.2cm]
		&  &  \scriptsize CVPR19 &  \scriptsize CVPR20  & \scriptsize ECCV20   & \scriptsize  NeurIPS20 & \scriptsize PAMI21 & \scriptsize CVPR21 & \scriptsize CVPR21  & \scriptsize ICCV21 &  \scriptsize PAMI2023 &   \multirow{2}{*}{Ours}\\
		[-0.2cm]
		& Metric & \scriptsize \cite{zhang2019csmg}  &  \scriptsize \cite{Zhang2020GCAGC}  & \scriptsize \cite{zhang2020gicd}  & \scriptsize \cite{Jin2020ICNet} & \scriptsize \cite{deng2021cosod3k} & \scriptsize \cite{zhang2021deepacg} & \scriptsize \cite{fan2021GCoNet} & \scriptsize \cite{zhang2021summarize} & \scriptsize \cite{zhu2023corp}  \\ \hline
		 & \cellg $\text{MAE}\downarrow$  &0.114 & 0.111   & 0.126  &0.148 & 0.106 &0.102 &0.105 &0.132 &  {0.121} &\textbf{0.099}\\ 
 		& $F_{\max}\uparrow$ &0.499 & {0.517} &  0.513   & 0.514 & 0.493 & 0.552&0.544 &0.548 &  {0.551} &\textbf{0.562}\\
		 & \cellg $S_{\alpha}\uparrow$ &0.627 & {0.666} & 0.658 & 0.657 & 0.612  &0.688 &0.673 &0.681 &  {0.686} &\textbf{0.712}\\
		\multirow{-4}{*}{\begin{sideways}CoCA\end{sideways}}
		& $E_{\xi}\uparrow$ &0.606 & 0.668 & {0.701} & 0.686 & 0.679  &- &0.739 &- & {0.715} &\textbf{0.728}\\
		\hline
	\end{tabular}
     \vspace{-11pt}
 
	\label{tab:cosod}
\end{table*}

\section{Experiments}
\label{sec:experiments}
\subsection{Datasets and Metrics}\label{sec:dataset-and-metric}
\vspace{-1pt}
To comprehensively evaluate GRSer's performance, apart from the proposed GRD, we also introduce RES and Co-SOD datasets as supplements.
For RES dataset (\eg, RefCOCO~\cite{yu2016modeling}, RefCOCO+~\cite{yu2016modeling}, and RefCOCOg~\cite{mao2016generation}),
 given that there exist some repeated sentences in different images, we reconstruct these datasets to the form of ``one sentence \vs a group of referred images", named as G-RefC, G-RefC+, and G-RefCg, with randomly sampled negative samples from other groups. 
 These re-built datasets have $8717$, $8020$, and $2451$ image groups, respectively, with positive to negative sample ratio of $1:1$ for both training and inference.
 In our experiment, GRD and re-built RES datasets are regarded as RES setting if the negative samples of the dataset are removed, otherwise it is the GRES setting.
 Besides, we use the CoCA~\cite{zhang2020gicd} dataset to evaluate our model's performance in Co-SOD task, where we take category names as expression inputs.

We adopt the metric of mean intersection-over-union (mIoU) for evaluating model's performance in RES setting with no negative samples included. When negative samples are introduced, their corresponding ground-truth masks are $0$ mask, where the originally defined mIoU is not valid (\ie, IoU $\equiv0$, for the negative sample). Therefore, we define an adapted metric $\overline{\text{mIoU}}$ to measure model performance on both segmentation accuracy and recognition ability for negative samples. Specifically, the idea of confusion matrix is adopted: for a true positive sample (TP), its $\overline{\text{IoU}}$ is calculated in the same way as the vanilla IoU; for a true negative sample (TN), its $\overline{\text{IoU}}$ is set to 1; for a false positive sample (FP) or false negative sample (FN), its $\overline{\text{IoU}}$ is set to 0. Then, the $\overline{\text{IoU}}$ value of all $m$ test samples are averaged to get the $\overline{\text{mIoU}}$, \ie, $\overline{\text{mIoU}}=\frac{1}{m}\sum_{i=1}^{m}\overline{\text{IoU}}_{i}$.
Besides, for Co-SOD task, common metrics of mean absolute error (MAE)\cite{cheng2013efficient}, maximum F-measure~\cite{borji2015SalObjBenchmark} ($F_{\max}$), S-measure~\cite{fan2017structure} ($S_{\alpha}$), and mean E-measure~\cite{Fan2018Em} ($E_{\xi}$) are adopted.

\subsection{Implementation Details}
The Transformer layers for visual encoding are initialized with classification weights pre-trained on ImageNet-22K from the Swin Transformer~\cite{liu2021swin}. The language encoder is the base BERT~\cite{devlin-etal-2019-bert} with $12$ layers and hidden size of $768$ (\ie, $C_{l}$), which is implemented from HuggingFace’s Transformer library~\cite{wolf2020transformers}. $C_{v}$ is set to $512$. Following~\cite{liu2021swin,yang2022lavt}, the AdamW optimizer~\cite{loshchilov2017decoupled} is adopted with weight decay of $0.01$. The initial learning rate is set to $0.00005$ with polynomial learning rate decay. The model is trained for 80 epochs with batch size of $4$. Images are resized to $416\times416$ and no data augmentations are employed. The size of input image group $N$ is set to $8$. The margin value $m$ is set to $1$ in triplet margin loss. 


\begin{table*}[t!]
\centering
\renewcommand{\arraystretch}{0.9}
\renewcommand{\tabcolsep}{2mm}
\caption{
\textbf{Ablation studies of ranking criteria} in heatmap hierarchizer on the G-RefCg and the proposed GRD datasets in the GRES setting. (*) indicates default choices of our model. The best results are marked in \textbf{bold}.
	}
     \vspace{-8pt}
 
\begin{tabular}{l|l|ccc|ccc}
\hline 
\multirow{2}{*}{Train}  & \multirow{2}{*}{Test} & \multicolumn{3}{c|}{G-RefCg}                                                                                            & \multicolumn{3}{c}{GRD}                                                                                          \\ 
                   &     & $\overline{\text{mIoU}}$ &$E_{\xi}$ &$R_{neg}$  & $\overline{\text{mIoU}}$ &$E_{\xi}$ &$R_{neg}$\\ 
                        \hline 
 \multirow{2}{*}{\textit{Random}} & \cellg  \textit{Random}         &68.92&0.554&89.12&50.73&0.475&75.38  \\
       & $R^{pos}+R^{neg}$  &      68.42&0.550&88.74&50.24&0.470&73.37   \\
      \hline
        \multirow{2}{*}{$R^{pos}+R^{neg}$ (*) } & \cellg \textit{Random} &67.79&0.548&88.23&49.83&0.468&73.28   \\
    & $R^{pos}+R^{neg}$ (*) &\textbf{75.32}&\textbf{0.572}&\textbf{95.25}&\textbf{57.12}&\textbf{0.515}&\textbf{81.09}   \\
      \hline
        $R^{pos}$ &\cellg  $R^{pos}$  &74.57&0.570&94.32&56.28&0.502&80.08   \\
          $R^{neg}$ & $R^{neg}$  &73.79&0.563&94.53&56.37&0.507 &80.23  \\
      \hline
\end{tabular}
     \vspace{-12pt}

\label{tab:abl_rank}
\end{table*}

\begin{table}[t!]
\centering
\renewcommand{\arraystretch}{1.1}
\renewcommand{\tabcolsep}{0.9mm}
\caption{
\textbf{Ablation studies of main designs} in our method on G-RefCg and GRD datasets in the GRES setting.
	}
     \vspace{-8pt}
 
\begin{tabular}{l|ccc|ccc}
\hline 
\multirow{2}{*}{}  & \multicolumn{3}{c|}{G-RefCg}                                                                                            & \multicolumn{3}{c}{GRD}                                                                                          \\ 
                        & $\overline{\text{mIoU}}$ &$E_{\xi}$ &$R_{neg}$  & $\overline{\text{mIoU}}$ &$E_{\xi}$ &$R_{neg}$\\ 
                        \hline 
\rowg       w/o. TQM   &66.28&0.525&85.33&47.62&0.463&70.29   \\
      w/o. HMapHier   &68.92&0.554&89.12&50.73&0.475&75.38  \\
\rowg       w/o. MirrorT &69.38&0.543&90.12&51.47&0.479&75.54   \\
      w/o. TriLoss    &30.37&0.493&0&23.14&0.435&0   \\
      \hline
      \rowg Full model   &\textbf{75.32}&\textbf{0.572}&\textbf{95.25}&\textbf{57.12}&\textbf{0.515}&\textbf{81.09}   \\
      \hline
\end{tabular}
     \vspace{-12pt}

\label{tab:abl_main}
\end{table}

\subsection{Comparison with SOTA Methods}
\noindent\textbf{Results on the GRES Setting.} 
In \tabref{tab:res}, we compare our GRSer with other RES methods on the re-built G-RefC, G-RefC+, G-RefCg, and our proposed GRD datasets. 
Specifically, negative samples are introduced to each image group for both training and inference (see \secref{sec:dataset-and-metric} for details), where the adapted metric $\overline{\text{mIoU}}$ is used.  
Compared methods are implemented following their original paradigms to input data in the form of ``one image \vs one expression". 
The ground-truth for the negative sample is set as $0$ mask.
Note that the proposed GRD dataset is only used for inference, and its corresponding train set is the combination of train sets from G-RefC, G-RefC+ and G-RefCg. 
It can be seen that our GRSer significantly outperforms other methods, and excels in recognition of negative samples, due to our designed triplet loss and mirror training strategy, which effectively optimize the multi-modal representation space. 

\noindent\textbf{Results on the RES Setting.} 
In \tabref{tab:res}, we present the results in the conventional RES setting, where mIoU metric is adopted. 
Here, no negative sample is included and all images in a group do contain target objects. 
Similarly, our GRSer outperforms all compared methods, particularly on the more difficult G-RefCg and GRD dataset (the given expressions are complex and hard to understand by models). It is the triphasic feature interations in TQM (\secref{sec:GCoQuery}) that help our model comprehend the complex semantics of the same object from different images in a group.

\noindent\textbf{Results on the Co-SOD Task.}
In \tabref{tab:cosod}, we further compare our GRSer with methods in the Co-SOD task on the CoCA dataset, where metrics including mean absolute error (MAE), maximum F-measure ($F_{\max}$), S-measure ($S_{\alpha}$), and mean E-measure ($E_{\xi}$) are adopted. Our model is trained on the combination of train sets from G-RefC, G-RefC+ and G-RefCg. The given category names of CoCA are regarded as expressions for grouped images during implementation. It can be seen that our method also achieves remarkable performances on this challenging real-world dataset.

\subsection{Ablation Studies}
\noindent\textbf{Triphasic Query Module (TQM).}
We remove the proposed TQM, and only a single language-activated heatmap is concatenated with visual features and then fed to the mask predictor. In \tabref{tab:abl_main}, the removal of TQM leads to a $\overline{\text{mIoU}}$ drop of $9.04\%$ and $9.50\%$ in G-RefCg and GRD, respectively, validating the effects of TQM. Besides, we try differnt image numbers in one group. In \tabref{tab:abl_group}, when increasing the group size $N$, model performances get better. 


\begin{table}[t!]
\centering
\renewcommand{\arraystretch}{1.1}
\renewcommand{\tabcolsep}{1.50mm}
\caption{
\textbf{Ablation studies of group size ($N$)} in TQM on G-RefCg and GRD datasets in the GRES setting. (*) indicates default choices of our model.
	}
     \vspace{-8pt}
 
\begin{tabular}{l|ccc|ccc}
\hline 
\multirow{2}{*}{}  & \multicolumn{3}{c|}{G-RefCg}                                                                                            & \multicolumn{3}{c}{GRD}                                                                                          \\ 
                        & $\overline{\text{mIoU}}$ &$E_{\xi}$ &$R_{neg}$  & $\overline{\text{mIoU}}$ &$E_{\xi}$ &$R_{neg}$\\ 
                        \hline 
\rowg       $N=1$   &66.28&0.525&85.33&47.62&0.463&70.29    \\
      $N=3$     &72.98&0.559&92.38&54.89&0.484&78.23   \\
\rowg       $N=5$     &74.26&0.567&94.45&56.01&0.502&80.92   \\
      $N=8$(*)   &\textbf{75.32}&\textbf{0.572}&\textbf{95.25}&\textbf{57.12}&\textbf{0.515}&\textbf{81.09}   \\   
      \hline
\end{tabular}
     \vspace{-12pt}
\label{tab:abl_group}
\end{table}
\noindent \textbf{Heatmap Hierarchizer (HMapHier).}
To explore the effects of the heatmap order in HMapHier, we experiment with different ranking criteria. 
As shown in \tabref{tab:abl_rank}, removing HMapHier (\ie, heatmap orders in both training and testing are random) results in the $\overline{\text{mIoU}}$ drops of $6.40\%$ and $6.39\%$ in G-RefCg and GRD, respectively. 
Besides, inconsistent ranking criteria in training and testing resulted in inferior performance.
Also, using the combination of positive rank $R^{pos}$ and negative rank $R^{neg}$ achieves the best results compared to using a single-source criterion.

\noindent \textbf{Mirror Training (MirrorT).}
In \tabref{tab:abl_main}, removing MirrorT leads to a $\overline{\text{mIoU}}$ drop of $5.94\%$ and $5.65\%$ in G-RefCg and GRD, respectively. 
This is because MirrorT plays a vital role in forcing model to comprehend the semantics contained in anti-expressions, helping our GRSer to be better aware of the image background and negative samples. 

\noindent \textbf{Triplet Margin Loss (TriLoss).}
\tabref{tab:abl_main} shows that TriLoss is critical for GRSer when negative samples are included. 
Without TriLoss, the recall of negative samples $R_{neg}$ falls to $0$ in both datasets, which means the model fails to recognize negative samples and output non-zero predicted masks for all images. 
TriLoss optimizes the multi-modal representation distances during training and constructs a well-distributed representation space that helps our model to distinguish between positive and negative samples.

\section{Conclusion}
In this work, we present a realistic multi-modal setting named Group-wise Referring Expression Segmentation (GRES), which relaxes the limitation of idealized setting in RES and extends it to a collection of related images. To facilitate this new setting, we introduce a challenging dataset named GRD, which effectively simulates the real-world scenarios by collecting images in a grouped manner and annotating both positive and negative samples thoroughly. Besides, a novel baseline method GRSer is proposed to explicitly capture the language-vision and vision-vision feature interactions for better comprehension of the target object. Extensive experiments show that our method achieves SOTA performances on GRES, RES, and Co-SOD.

\balance
{\small
\bibliographystyle{ieee_fullname}
\bibliography{main}
}

\end{document}